\documentclass[letterpaper]{article}

\usepackage{aaai21}
\usepackage{times}
\usepackage{helvet}
\usepackage{courier}
\usepackage[hyphens]{url}
\usepackage{graphicx}
\usepackage{amssymb}
\usepackage{amsmath}
\usepackage{natbib}
\usepackage{caption}
\usepackage{macros}
\usepackage{algorithm}
\usepackage{algpseudocode}
\usepackage{paralist}
\usepackage{colonequals}
\usepackage{xspace}
\usepackage[frozencache,cachedir=minted]{minted}

\def\x{\hspace{-5pt}}

\title{Neural-Symbolic Integration: A Compositional Perspective}
\author{
Efthymia Tsamoura,\textsuperscript{\rm 1}
Loizos Michael \textsuperscript{\rm 2} \\
}
\affiliations{
\textsuperscript{\rm 1} Samsung Research \\
\textsuperscript{\rm 2} Open University of Cyprus \& \\
Research Center on Interactive Media, \\
Smart Systems, and Emerging Technologies \\
efi.tsamoura@samsung.com, loizos@ouc.ac.cy}

\begin{document}

\maketitle

\begin{abstract}
Despite significant progress in the development of neural-symbolic frameworks, the question of how to integrate a neural and a symbolic system in a \emph{compositional} manner remains open. Our work seeks to fill this gap by treating these two systems as black boxes to be integrated as modules into a single architecture, without making assumptions on their internal structure and semantics. Instead, we expect only that each module exposes certain methods for accessing the functions that the module implements: the symbolic module exposes a deduction method for computing the function's output on a given input, and an abduction method for computing the function's inputs for a given output; the neural module exposes a deduction method for computing the function's output on a given input, and an induction method for updating the function given input-output training instances. We are, then, able to show that a symbolic module --- with any choice for syntax and semantics, as long as the deduction and abduction methods are exposed --- can be cleanly integrated with a neural module, and facilitate the latter's efficient training, achieving empirical performance that exceeds that of previous work.
\end{abstract}

\section{Introduction} %
\label{Section: Introduction}

Neural-symbolic frameworks \cite{DBLP:books/daglib/0007534,rocktaschel2017ntp,wang2019satNet} vow to bring a new computational paradigm in which symbolic systems can tolerate noisy or unstructured data, and neural systems can learn with fewer data and offer interpretable outcomes.
The potential of integrating a symbolic, typically logic-based, module on top of a neural one has been well-demonstrated in semi-supervised learning \cite{DBLP:journals/corr/DonadelloSG17,DBLP:journals/corr/abs-1901-04195,DBLP:journals/corr/SerafiniG16,aa21302d28c4436aac754a7ac90bfacf}, program induction \cite{kalyan2018neuralguided,DBLP:conf/iclr/ParisottoMS0ZK17}, and open question answering \cite{sun2018openQA} settings. In these cases, the training of the neural module is regulated by the logic theory (and its integrity constraints or other constructs), which is far from straightforward since logical inference cannot be, in general, captured via a differentiable function. 

To accommodate the integration of neural modules with logical theories, the majority of neural-symbolic frameworks restrict the type of the theories (e.g., to non-recursive or acyclic propositional ones), and they either translate them into neural networks \cite{DBLP:books/daglib/0007534,HSK1999,TOWELL1994119}, or they replace logical computations by differentiable functions \cite{pmlr-v70-bosnjak17a,pmlr-v70-gaunt17a}. A second line of work abandons the use of classical logic altogether and adopts theories whose interpretations take continuous values, such as fuzzy logic \cite{DBLP:journals/corr/DonadelloSG17,DBLP:journals/corr/abs-1901-04195,DBLP:journals/corr/SerafiniG16,DBLP:journals/corr/SourekAZK15,aa21302d28c4436aac754a7ac90bfacf}, or probabilistic logic \cite{DBLP:journals/corr/abs-1907-08194}, which can support the uniform application of backpropagation on both the symbolic and the neural module. 

We consider the problem of integrating a symbolic module that computes a function $s(\cdot)$ on top of a neural module that computes a function $n(\cdot)$, so that together the two modules implement the composition $s \circ n$. We argue that this integration can be done fully compositionally, without the need to revamp the syntax or semantics of either module. 

We borrow two well-known notions from mathematical logic to establish the interface that should be provided by the symbolic module to reach a transparent and ``non-intrusive'' integration: \emph{deduction}, or forward inference, and \emph{abduction}, through which one computes (i.e., abduces) the inputs to the symbolic module that would deduce a given output.     

While abduction has been used in the past as the means to train a neural module feeding into a symbolic module \cite{ABL}, there are two key differences between our framework and prior art, over and above our high-level contribution in setting the basis for compositionality. 
The first difference is on the abduced inputs that are used to train the neural module. Our basic framework makes use of \emph{all} such abduced inputs, while prior art restricts its attention on one of them. As also supported by the empirical evidence that we offer in this work, this restriction causes the learning process to suffer: learning is led to fall into local minima since the single abduced input offers lopsided feedback to the learning process, training faces weaker supervision signals due to the \emph{loss} of the semantic constraints among the different abduced inputs, and the learning process becomes vulnerable to random supervision signals on those parts of the single abuced input that are forced to take values when they should have semantically been treated as irrelevant. 

The second difference is on the training process itself. Prior art uses an ad-hoc training procedure which requires training of the neural module multiple times for the same training sample. That training approach is not only computationally expensive,  but it is also difficult to customize on different scenarios. Instead, our framework provides the means to control the training process in a customized manner by delegating to the symbolic module the encoding of any domain-specific training choices. In particular, there exist cases where one would wish to have the neural predictions guide the choice of abduced inputs --- presumably the problem that also motivates prior art. We show that such \emph{neural-guided abduction} can be done easily as an extension of our basic framework, by encoding in the symbolic module the knowledge of which abduced inputs are to be used for training, using declarative or procedural techniques to resolve any inconsistencies and to rank the abduced inputs in terms of compatibility with the current neural predictions.

Beyond the plugging in of theories with any semantics and syntax, and beyond the already-mentioned support for neural-guided abduction, the clean take of our proposed compositional architecture easily extends to support other features found in past works, including \emph{program induction} and \emph{domain-wide constraints}. To our knowledge, a uniform handling of all these features is not present in past works.

We empirically evaluate --- in what we believe to be a more comprehensive manner than typically found in the relevant literature --- the performance of our framework against three frameworks that share the same goals with ours: $\deepproblog$ \cite{DBLP:journals/corr/abs-1907-08194}, $\neurasp$ \cite{neurasp}, and $\abl$ \cite{ABL}. We demonstrate the superior performance of our framework both in terms of training efficiency and accuracy over a wide range of scenarios showing the features described above.

\section{Preliminaries} %
\label{Section: Preliminaries}

For concreteness of exposition, and without excluding other syntax and semantics, we assume that the symbolic component encodes a logic theory using the standard syntax found in the abductive logic programming literature \citep{Kakas2017}.

\begin{figure}[t]
    \centering
    \includegraphics[width=\columnwidth]{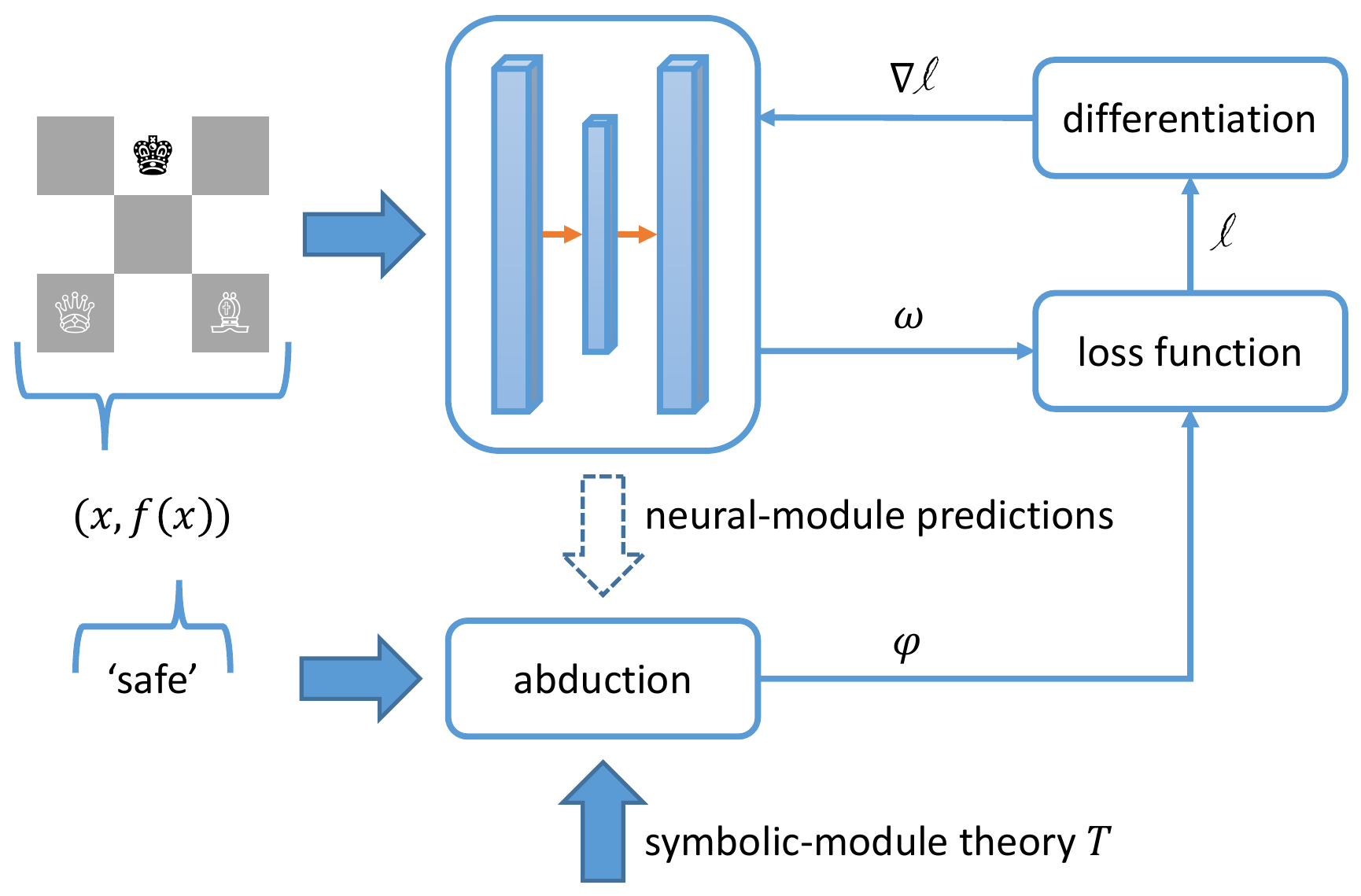}
\tiny
\begin{minted}{prolog}
safe :- placed(Z1), movable(Z1).
draw :- placed(Z1), \+attacked(Z1), \+movable(Z1).
mate :- placed(Z1), attacked(Z1), \+movable(Z1).

placed(Z1) :- pos(Z1), at(b(k),Z1), pos(Z2), pos(Z3), Z2\=Z3,
              piece(w(P2)), at(w(P2),Z2), piece(w(P3)), at(w(P3),Z3).

movable(Z1) :- pos(Z2), reached(Z2,k,Z1), \+attacked(Z2).
attacked(Z2) :- pos(Z3), piece(w(P)), at(w(P),Z3), reached(Z2,P,Z3).

reached((X,Y),k,(PX,PY)) :- abs(X,PX,DX), 1>=DX, abs(Y,PY,DY),
                            1>=DY, sum(DX,DY,S), 0<S.
reached((X,Y),q,(PX,PY)) :- reached((X,Y),r,(PX,PY)).
reached((X,Y),q,(PX,PY)) :- reached((X,Y),b,(PX,PY)).
...

ic :- piece(P), at(P,Z1), at(P,Z2), Z1\=Z2.
ic :- piece(P1), piece(P2), at(P1,Z), at(P2,Z), P1\=P2.
ic :- at(b(k),Z1), at(w(k),Z2), reached(Z1,k,Z2).
ic :- piece(b(P1)), at(b(P1),Z1), piece(b(P2)), at(b(P2),Z2), Z1\=Z2.
ic :- piece(w(P1)), at(w(P1),Z1), piece(w(P2)), at(w(P2),Z2),
      piece(w(P3)), at(w(P3),Z3), Z1\=Z2, Z2\=Z3, Z3\=Z1.
\end{minted}
\caption{Snippet of a theory for an example chess domain being used to train a neural module through abduction.}
\label{figure:architecture}
\end{figure}

As typical in logic programming, the language comprises a set of relational \notion{predicates} that hold over \notion{variables} or \notion{constants}. An \notion{atom} is a predicate with its arguments. A \notion{formula} is defined as a logical expression over atoms, using the logical connectors of Prolog, e.g., conjunction, disjunction, negation. A \notion{theory} is a collection of such formulas. Figure~\ref{figure:architecture} shows a theory for determining the status of the game of a certain variant of chess played on a $3 \times 3$ board with three pieces: a black king, and two white pieces of different types.

As far as our proposed architecture is concerned, the precise syntax and semantics of the theory are inconsequential. We will, therefore, not delve into a detailed analysis of the aforementioned theory $T$, except as needed to highlight certain features. What is of only importance is that $T$ is accompanied by an entailment operator $\models$ that allows exposing: a deduction method $\deduce$ that takes as input a set of atoms $A$ and produces a set of atoms $O = \deduce(T,A)$ such that $T \cup A \models O$; an abduction method $\abduce$ that takes as input a set of atoms $O$ and produces a set (out of possibly many) of atoms $A \in \abduce(T,O)$ such that $T \cup A \models O$.

As part of exposing a method one needs to define its input and output spaces. We will assume that $A \subseteq \As$ and call $\As$ the set of \notion{symbolic inputs} or \notion{abducibles}; and that $O \subseteq \Os$ and call $\Os$ the set of \notion{symbolic outputs} or \notion{outcomes}. We will also assume that atoms in $\As$ and $\Os$ are grounded and disjoint. When convenient, we will represent a subset of atoms as a formula: the conjunction of the subset's members. An \notion{abductive proof} for a given outcome $O \subseteq \Os$ is any formula $A \in \abduce(T,O)$. Observe that for any fixed outcome there might exist zero, one, or multiple abductive proofs.

\begin{example}\label{example:chess1}
In our example chess domain, the set $\As$ of abducibles comprises all atoms of the form ${\at(P,(X,Y))}$,  corresponding to the concept of a chess piece of type $P$ being on the chess board at coordinates ${(X,Y)}$; $P$ takes one of the values in $\set{\bck, \wck, \wcq, \wcr, \wcb, \wcn, \wcp}$, where $\wc(\cdot)$ and $\bc(\cdot)$ stand for white or black pieces, and $k$, $q$, $r$, $b$, $n$, and $p$ denote the king,  queen, rook, bishop, knight and pawn, respectively; each of $X$ and $Y$ take one of the values in $\set{1,2,3}$. The set $\Os$ of outcomes is equal to $\set{\safe, \draw, \mate}$, corresponding to the concepts that the black king has a valid move, is stalemated, or is mated.

Deduction receives as input a subset of $\As$ that describes the state of the chess board, and produces as output a (singleton) subset of $\Os$ on the status of the black king. Conversely, abduction receives as input a (singleton) subset of $\Os$ that describes the desired status of the black king, and produces as output subsets of $\As$, each describing a state of the chess board where the black king has the desired status.
\end{example}

A theory may be extended with \notion{integrity constraints}, special formulas that restrict the possible inferences that can be drawn when applying the methods of deduction and abduction, by constraining which subsets of $\As$ are considered acceptable. A subset $A \subseteq \As$ violates the integrity constraints if and only if $\deduce(T,A)$ is a special symbol $\bot \not\in \Os$. Analogously, a subset $A \subseteq \As$ violates the integrity constraints if and only if $A \not\in \abduce(T,O)$ for each subset $O \subseteq \Os$. Thus, integrity constraints in a theory need to be respected by every abductive proof for each outcome $O \subseteq \Os$.

\begin{example}
\label{example:chess2}
In our example chess domain, the integrity constraints are encoded as rules with an \texttt{ic} head. The five integrity constraints in Figure~\ref{figure:architecture} capture, in order, the following requirements: the same piece type is not at more than one position; no two pieces are at the same position; the black and white kings are not attacking each other; there is at most one black piece on the chess board; there are at most two white pieces on the chess board. The requirement for the existence of at least one black king and at least two white pieces is captured through the formula with the \texttt{placement} head. If the set $\As$ of abducibles is extended to include all atoms of the form $\ec((X,Y))$ to denote explicitly the coordinates of the board cells that are empty, then additional integrity constraints and formulas can be added in the theory to ensure that no piece can be placed at an empty cell, an that every non-empty cell should hold a piece.
\end{example}

\section{Framework} %
\label{Section: Framework}

We consider a neural-symbolic system built by composing a neural module feeding into a symbolic module.

\subsection{Module Compositionality} %
\label{Subsection: Module Compositionality}

We let $\Xs$ and $\Omega = [0,1]^k$ be, respectively, the space of possible inputs and the space of possible outputs of the neural module. At any given training iteration $t$, the neural module effectively implements a function $n_t : \Xs \rightarrow \Omega$. For notational simplicity, we will overload the use of the symbol $n_t$ to denote both the function and the underlying neural network itself. We assume that there is a \notion{translator} function $r$ that maps each $\omega \in \Omega$ to a set of abducibles $r(\omega) \in \As$.

Given a symbolic module with a theory $T$, the \notion{end-to-end reasoning} of the neural-symbolic system at iteration $t$ is the process that maps an input in $\Xs$ to an outcome subset of $\Os$ as follows: the system receives an input $x \in \Xs$; the neural module computes the vector $\omega = n_t(x)$; the translator maps $\omega$ to the abducibles $A = r(\omega) \subseteq \As$; the symbolic module computes the outcome $O = \deduce(T,A) \subseteq \Os$. Thus, \emph{inference} in our framework proceeds by running the inference mechanism of the symbolic module over the inferences of the neural module on a given neural input. To simplify our notation, and when there is no confusion, we will often write $h^T_t(x)$ to mean $\deduce(T,r(n_t(x)))$ for every $x \in \Xs$.

\begin{example}
\label{example:chess3}
In our example chess domain, consider a neural module $n_t$ that receives as input $x \in \Xs$ a $3 \times 3$ grid of images representing a chess board. The neural module outputs a vector $\omega = n_t(x) \in \Omega^k$ that corresponds to what the neural module predicts. One possible implementation is for the neural module to have eight output nodes for each cell at coordinates $(X,Y)$ of the chess board (hence, $k = 8 \times 9$). These eight output nodes represent, respectively, whether their associated cell includes no piece, the black king, the white king, the white queen, the white rook, the white bishop, the white knight, or the white pawn. $\omega$ assigns, effectively, confidence values on each of these predictions for each cell.

The translator function $r$ could simply turn $\omega$ into a set of abducibles $A$ by considering for each cell the most confident prediction and including the corresponding atom in $A$. Thus, if the first eight components of $\omega$, which correspond to predictions for cell $(1,1)$, where such the third value was the maximum one, then $A$ would include $\at(\wck,(1,1))$.

$A$ is provided as input to the symbolic component, which deduces whether the chess board is in a safe, draw, or mate state (or in $\bot$ in case $A$ violates the integrity constraints).
\end{example}

In certain cases, the input $x \in \Xs$ to a neural-symbolic system might be associated with explicit input-specific knowledge provided in the form of a symbolic formula $\overline{x}$. This side-information does not go through the usual pipeline as $x$, but can be readily accommodated by extending the theory $T$ to include it, and by computing $\deduce(T \cup \set{\overline{x}},r(n_t(x)))$ instead. Our compositional perspective affords us to remain agnostic, at the architecture level, on how side-information will be dealt with by the symbolic module (e.g., as integrity constraints or as weak preferences), and puts the burden on the theory itself to make this domain-specific determination.

\subsection{Neural-Module Learning} %
\label{Subsection: Neural-Module Learning}

As in standard supervised learning, consider a set of labeled samples of the form ${\set{\tuple{x_j,f(x_j)}}_j}$, with $f$ being the target function that we wish to learn, $x_j$ corresponding to the features of the sample, and $f(x_j)$ being the label of the sample.

In the context of our neural-symbolic architecture, learning seeks to identify, after $t$ iterations over a training subset of labeled samples, a hypothesis function $h^T_t(\cdot)$ that sufficiently approximates the target function $f(\cdot)$ on a testing subset of labeled samples. Given a fixed theory $T$ for the symbolic module, the only part of the hypothesis function $h^T_t(\cdot) = \deduce(T,r(n_t(\cdot)))$ that remains to be learned is the function $n_t$ implemented by the neural module.

We put forward Algorithm~\ref{algorithm:train} to achieve this goal. In line with our compositional treatment, the algorithm does not delve into the internals of the neural and the symbolic module, but accesses them only through the methods that they expose: inference and backpropagation for the neural module; deduction and abduction for the symbolic module.

\begin{algorithm}[t]
\caption{\textsc{train}($x, f(x), n_t$) $\rightarrow$ $n_{t+1}$}
\label{algorithm:train}
\begin{algorithmic}[1]

\State ${\omega \defeq n_t(x)}$ \label{algorithm:train:neural}

\State $\varphi \defeq \bigvee \abduce(T,f(x))$ \Comment{basic form \emph{or}} $\varphi \defeq \bigvee \abduce(T \cup r(\omega),f(x))$ \Comment{NGA form}

\label{algorithm:train:abduction}	
		
\State $\ell \defeq \mathsf{loss}(\varphi, r, \omega)$ \Comment{using WMC}

\State $n_{t+1} \defeq \mathsf{backpropagate}(n_t, \bigtriangledown \ell)$

\State \textbf{return} $n_{t+1}$
\end{algorithmic}
\end{algorithm}

The algorithm considers the label $f(x)$ of a given sample, viewed as a (typically singleton) subset of $\Os$, and abduces \emph{all} abductive proofs $A \in \abduce(T,f(x)) \subseteq \As$. Taking the disjunction of all abductive proofs, the algorithm computes the \notion{abductive feedback} formula $\varphi$ that captures all the acceptable outputs of the neural module that would lead, through the theory T, the system to correctly infer $f(x)$.

The abductive feedback acts as a supervision signal for the neural module. Combining that signal with the actual output $\omega$ of the neural module (through the use of the translator function $r$), we can compute the loss of the neural module. \emph{Critically, the resulting loss function is differentiable, even if the theory $T$ of the symbolic module is not!} By differentiating the loss function we can use backpropagation to update the neural module to implement function $n_{t+1}$.

Rather than requiring for the theory to be differentiable, as done in certain past works \cite{DBLP:journals/corr/DonadelloSG17,DBLP:journals/corr/abs-1901-04195,DBLP:journals/corr/SerafiniG16,DBLP:journals/corr/SourekAZK15,aa21302d28c4436aac754a7ac90bfacf,DBLP:journals/corr/abs-1907-08194}, the use of abduction for neural-symbolic integration poses no a priori constraints on the form of the theory, but proceeds to extract its ``essence'' in a differentiable form, albeit in an outcome-specific manner. Fortuitously, the space of possible outcomes is usually considerably restricted, which readily allows the caching of the abductive proofs, or even their precomputation prior to the training phase. Put differently, the use of abduction allows replacing any arbitrary theory $T$ by the set of its abductive feedbacks $\set{\varphi_O \mid \varphi_O = \bigvee \abduce(T,O), O \subseteq \Os}$.

\begin{example}
\label{example:chess4}
In our example chess domain, consider a training sample $(x,f(x))$, where $x$ is a $3 \times 3$ grid of images representing a chess board with a a white queen at cell $(1,1)$, a white bishop at cell $(3,1)$, and a black king at cell $(2,3)$, and $f(x)$ labels the chess board as being in a safe state. Starting from the label, we compute the abductive feedback $\ldots \vee
[\at(\wcq,(1,1)) \wedge \at(\wcb,(3,1)) \wedge \at(\bck,(2,3)) \wedge \ldots \wedge \ec((3,3))] \vee
[\at(\wcr,(1,1)) \wedge \at(\wcn,(3,1)) \wedge \at(\bck,(2,3)) \wedge \ldots \wedge \ec((3,3))] \vee
[\at(\bck,(1,1)) \wedge \at(\wcp,(3,1)) \wedge \at(\wcr,(2,3)) \wedge \ldots \wedge \ec((3,3))] \vee
[\at(\wcp,(1,1)) \wedge \at(\wcn,(2,2)) \wedge \at(\bck,(2,3)) \wedge \ldots \wedge \ec((3,3))] \vee \ldots$.
Among the shown disjuncts, the first one represents the input chess board, the next two represent
chess boards that are safe and have pieces only at cells $(1,1)$, $(3,1)$ and $(2,3)$,
and the last represents a chess board that is safe, but has pieces at cells
$(1,1)$, $(2,2)$ and $(2,3)$.
\end{example}

\subsection{Neural-Guided Abduction} %
\label{Subsection: Neural-Guided Abduction}

Although computing the entire abductive feedback is generally the appropriate choice of action, there might exist circumstances where it might be beneficial to prune some of its parts. Caution should, however, be exercised, as pruning might end up removing the part of the abductive feedback that corresponds to the true state of affairs (cf.\ Example~\ref{example:chess4}), and might, thus or otherwise, misdirect the learning process.

One case worth considering is \notion{neural-guided abduction} (NGA), where the prediction of the neural module is used as a focus point, and only abductive proofs that are proximal perturbations of that point find their way into the abductive feedback. What counts as a perturbation, how proximity is determined, and other such considerations are ultimately domain-specific, and are not specified by the framework.

\begin{example}
\label{example:chess5}
In our example chess domain, consider a neural module that is highly confident in distinguishing empty from non-empty cells, but less confident in determining the exact types of the pieces in the non-empty cells. Consider, further, a particular training sample ${\tuple{x,f(x)}}$ on which the neural component identifies the non-empty cells as being ${(1,1)}$, ${(3,1)}$, and ${(2,3)}$. It is then natural for the symbolic module to attempt to utilize the predictions of the neural module to prune and focus the abductive feedback that it will provide for the further training of the neural module.

If, for example, $f(x)$ labels the chess board as being in a safe state, then the abductive feedback will exclude the last disjunct from Example~\ref{example:chess4}, since it represents a chess board with pieces at cells other than ${(1,1)}$, ${(3,1)}$, and ${(2,3)}$, and will maintain the first three disjuncts as they respect the neural predictions in terms of the positions of the three pieces.
\end{example}

To support neural-guided abduction, we must, first, establish a communication channel between the neural module and the abduction mechanism, in order for the neural module to provide its predictions to the abduction mechanism.

Our proposed architecture can seamlessly implement this communication channel by treating the communicated information as input-specific knowledge. Given, therefore, a training sample $(x,f(x))$, we can simply call the abduction method not by providing only the theory $T$ and the outcome $f(x)$ as inputs, but by first extending the theory $T$ with the neural predictions $\omega = n_t(x)$ as translated by the translator function $r$. Thus, the abductive feedback in Algorithm~\ref{algorithm:train} is now computed as $\varphi \defeq \bigvee \abduce(T \cup r(\omega),f(x))$.

As we have already mentioned, the treatment of this side-information is not determined by the framework, but is left to the theory itself. Although the side-information might, in some domains, provide confident predictions that could act as hard constraints for the theory (cf.\ Example~\ref{example:chess5}), our treatment allows also the handling of domains where the side-information might be noisy, incorrect, or even in direct violation of the existing integrity constraints of the theory.

Such neural predictions might still offer some useful guidance to the abduction process. Depending on the syntactic and semantic expressivity of the symbolic module, the theory can provide a declarative or a procedural way to resolve the inconsistencies that arise in a domain-specific manner.

\begin{example}
\label{example:chess6}
In our example chess domain, consider a particular training sample ${\tuple{x,f(x)}}$ on which the prediction of the neural module, as translated by the translator into symbolic inputs, corresponds to the subset $\{\at(\wcq,(1,1))$, $\at(\wcb,(3,1))$, $\at(\bck,(2,3))$, $\ldots$, $\ec((3,3))\}$.

Assume, first, that $f(x)$ labels the chess board as being in a safe state. Then, there exists exactly one abductive proof that matches the neural prediction perfectly. As this corresponds to a zero-cost perturbation of the neural prediction, only it ends up in the abductive feedback. As a result, the neural module ends up reinforcing exactly what it predicted.

Assume, now, that $f(x)$ labels the chess board as being in a draw state. Then, there is no abductive proof that matches the neural prediction perfectly. Rather, there is an abductive proof $[\at(\wcq,(1,1)) \wedge \at(\wcr,(3,1)) \wedge \at(\bck,(2,3)) \wedge \ldots \wedge \ec((3,3))]$ that differs from the neural prediction only in changing the type of an already predicted white piece, while maintaining its position, and also maintaining the types and positions of the other two pieces. This abductive proof could be evaluated to have a minimal-cost among the perturbations of the neural prediction, and only it ends up in the abductive feedback. As a result, the neural module ends up reinforcing parts of what it sees, while helping revise locally one of its mistakes (perhaps because it is still unable to fully differentiate between rooks and bishops).

Assume, finally, that $f(x)$ labels the chess board as being in a mate state. Then, there is no abductive proof that matches the neural prediction perfectly. In fact, there are no abductive proofs that respect the positions of the pieces as predicted by the neural module. Abduction will then seek to identify perturbations that, if possible, move a single piece with respect to the predicted ones, or move and change the type of a single piece, etc., that would respect the label $f(x)$. Depending on how one costs the various perturbations, one or more abductive proofs can be evaluated to have minimal-cost, and all those will end up in the abductive proof.
\end{example}

\section{Evaluation} %
\label{Section: Evaluation}

\begin{table*}[t]
\centering
\resizebox{0.75\textwidth}{!}{%
 \begin{tabular}{ |l c c c c c c| }
 \hline
 & $\add$ & $\operator$ & $\apply$ & $\dba(5)$ & $\mat(3)$ & $\mat(5)$ \\
 accur \% \hfill \textsc{NLog} & $91.7 \pm 0.7$ &  $90.8 \pm 0.8$ & ${100 \pm 0}$ & ${95.0 \pm 0.2}$ & ${95.0 \pm 1.2}$ & ${92.2 \pm 0.9}$ \\
 accur \% \hfill \textsc{DLog} & $88.4 \pm 2.5$ &  $86.9 \pm 1.0$ & ${100 \pm 0}$ & ${95.6 \pm 1.8}$ & ${93.4 \pm 1.4}$ & timeout \\ 
 accur \% \hfill \textsc{ABL}  & ${75.5 \pm 34}$ & timeout & ${88.9 \pm 13.1}$ & ${79 \pm 12.8}$ & ${69.7 \pm 6.2}$ & ${6.1 \pm 2.8}$ \\ 
 accur \% \hfill \textsc{NASP} & ${89.5 \pm 1.8}$ & timeout & ${76.5 \pm 0.1}$ & ${94.8 \pm 1.8}$ & ${27.5 \pm 34}$ & ${18.2 \pm 33.5}$ \\ \hline
 
 time (s) \hfill \textsc{NLog} & $531 \pm 12$ & ${565 \pm 36}$ & $228 \pm 11$ & ${307 \pm 51}$ & ${472 \pm 15}$ & ${900 \pm 71}$ \\
 time (s) \hfill \textsc{DLog} & ${1035 \pm 71}$ & ${8982 \pm 69}$ & ${586 \pm 9}$ & ${4203 \pm 8}$ & $1649 \pm 301$ & timeout \\
 time (s) \hfill \textsc{ABL}  & ${1524 \pm 100}$ & timeout & ${1668 \pm 30}$ &  ${1904 \pm 92}$ & ${1903 \pm 17}$ & ${2440 \pm 13}$ \\
 time (s) \hfill \textsc{NASP} & ${356 \pm 4}$  & timeout &  ${454 \pm 652}$ & ${193 \pm 2}$ & ${125 \pm 6}$ & ${217 \pm 3}$ \\
 \hline
\end{tabular}
}\\[5pt]
\resizebox{0.9\textwidth}{!}{%
 \begin{tabular}{ |l c c c c c c c| }
 \hline
 & $\pat(4)$ & $\pat(6)$ & $\memberof(3)$ & $\memberof(5)$ & $\minichessBSV(3)$ & $\minichessISK(3)$ & $\minichessNGA(3)$ \\
 accur \% \hfill \textsc{NLog} & ${97.4 \pm 1.4}$ & ${97.2 \pm 1.1}$ & ${96.9 \pm 0.4}$ & ${95.4 \pm 1.2}$ & ${94.1 \pm 0.8}$ & ${93.9 \pm 1.0}$ & ${92.7 \pm 1.6}$ \\
 accur \% \hfill \textsc{DLog} & timeout & timeout & ${96.3 \pm 0.3}$ & timeout & n/a & n/a & n/a \\ 
 accur \% \hfill \textsc{ABL}  & timeout & timeout & ${55.3 \pm 3.9}$ & ${49.0 \pm 0.1}$ & ${0.3 \pm 0.2}$ & ${44.3 \pm 7.1}$ & n/a \\
 accur \% \hfill \textsc{NASP} & timeout & timeout & ${94.8 \pm 1.3}$ & timeout &  timeout & ${19.7 \pm 6.3}$ & n/a \\ \hline
 
 time (s) \hfill \textsc{NLog} & ${958 \pm 89}$ & ${2576 \pm 14}$ & ${333 \pm 23}$ & ${408 \pm 18}$ & ${3576 \pm 28}$ & ${964 \pm 15}$ & ${2189 \pm 86}$ \\
 time (s) \hfill \textsc{DLog} & timeout & timeout & $2218 \pm 211$ & timeout & n/a & n/a & n/a \\
 time (s) \hfill \textsc{ABL}  & timeout & timeout & ${1392 \pm 8}$ & ${1862 \pm 28}$ & ${9436 \pm 169}$ & ${7527 \pm 322}$ & n/a \\
 time (s) \hfill \textsc{NASP} & timeout & timeout & ${325 \pm 3}$ & timeout &  timeout & ${787 \pm 307}$ & n/a \\
 \hline
\end{tabular}
}
\caption{Empirical results. \textsc{NLog} stands for $\ours$, \textsc{DLog} for $\deepproblog$ and \textsc{NASP} for $\neurasp$.}
\label{Table: Empirical Results}
\end{table*}

\begin{figure*}[t]
\centering
\begin{tabular}{ccccc}
\x
& \x\x $\add$
& \x\x $\operator$
& \x\x $\apply$
& \x\x $\dba(n)$ \\[-0.25ex]
\x \rotatebox{90}{\hspace{0.2cm}{\footnotesize accuracy \% / iter.}} 	 	
& \x\x \includegraphics[width=0.24\textwidth]{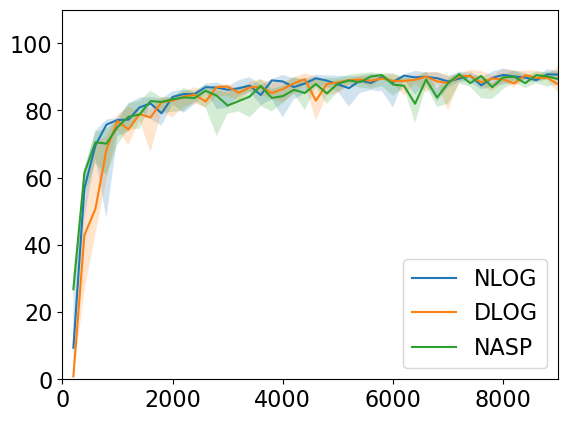} 
& \x\x \includegraphics[width=0.24\textwidth]{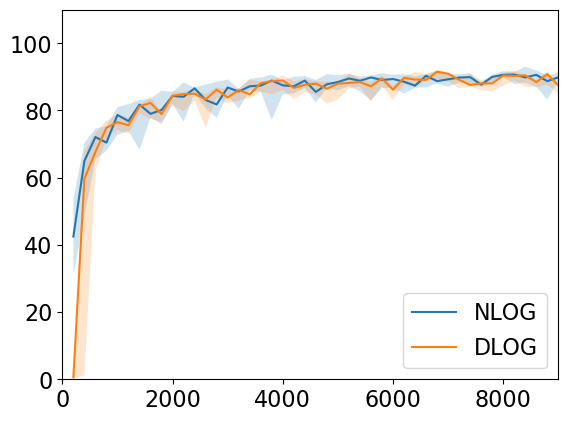}
& \x\x \includegraphics[width=0.24\textwidth]{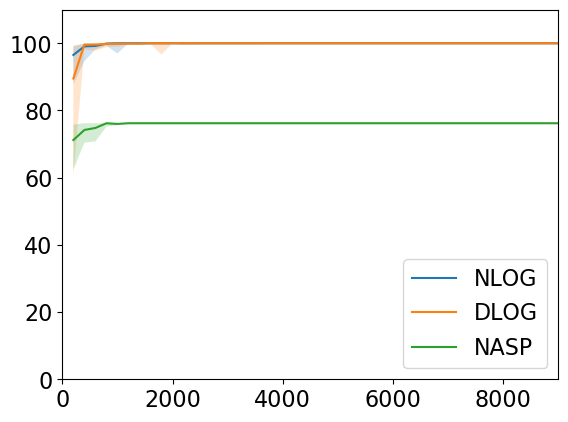}
& \x\x \includegraphics[width=0.24\textwidth]{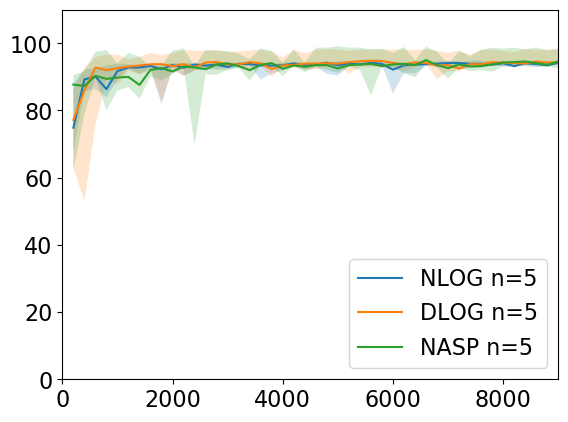} \\
\x
& \x\x $\mat(3)$
& \x\x $\mat(5)$
& \x\x $\pat$
& \x\x $\memberof(n)$ \\[-0.25ex]
\x \rotatebox{90}{\hspace{0.2cm}{\footnotesize accuracy \% / iter.}} 	 	
& \x\x \includegraphics[width=0.24\textwidth]{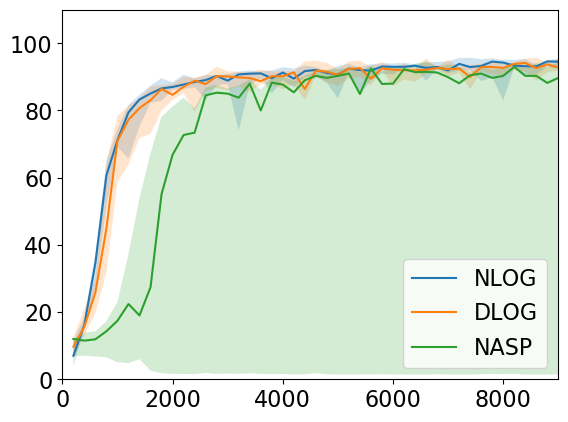}
& \x\x \includegraphics[width=0.24\textwidth]{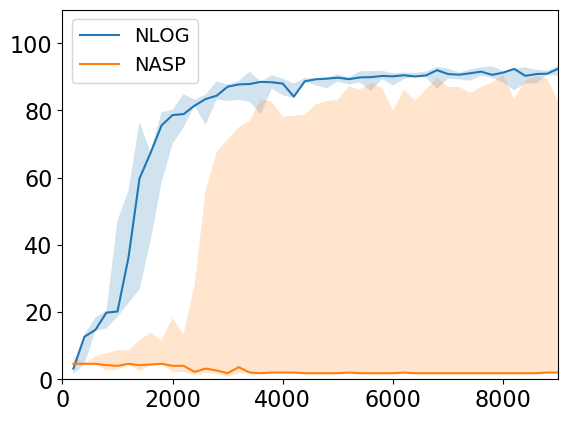}
& \x\x \includegraphics[width=0.24\textwidth]{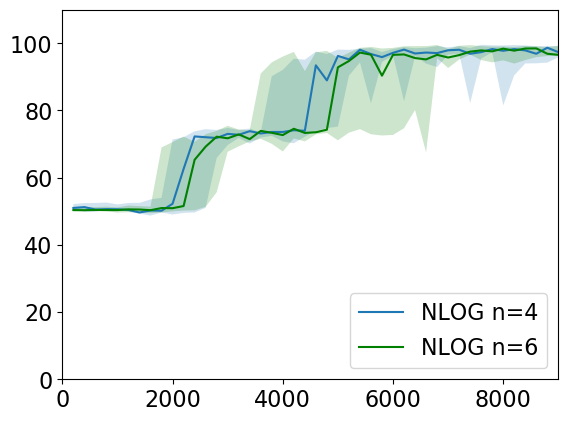} 
& \x\x \includegraphics[width=0.24\textwidth]{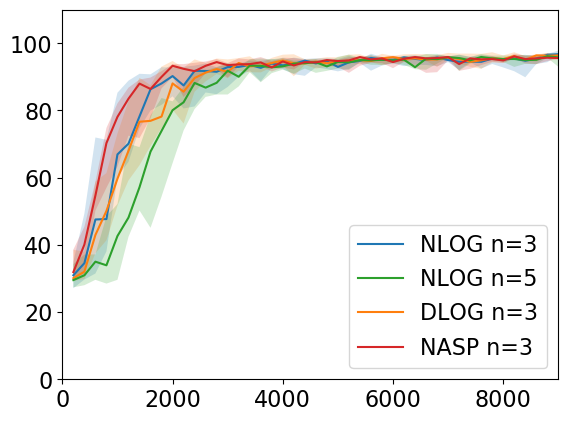} 
\end{tabular}\\[5pt]

\caption{Empirical results for $\ours$, $\deepproblog$, and $\neurasp$. Solid lines and lightly-colored areas show, respectively, the average behavior and the variability of the behavior across different repetitions.}
\label{Figure: Empirical Results 1}
\end{figure*}

\begin{figure*}[t]
\centering
\begin{tabular}{ccccc}
\x
& \x\x $\add$
& \x\x $\apply$
& \x\x $\dba(n)$
& \x\x $\mat(3)$ \\[-0.25ex]
\x \rotatebox{90}{\hspace{0.2cm}{\footnotesize accuracy \% / iter.}} 	 	
& \x\x \includegraphics[width=0.24\textwidth]{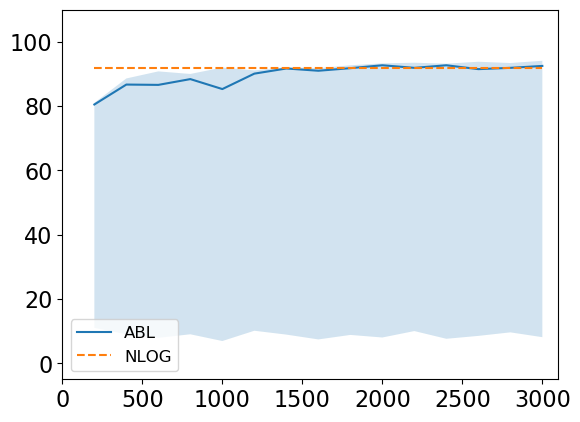} 
& \x\x \includegraphics[width=0.24\textwidth]{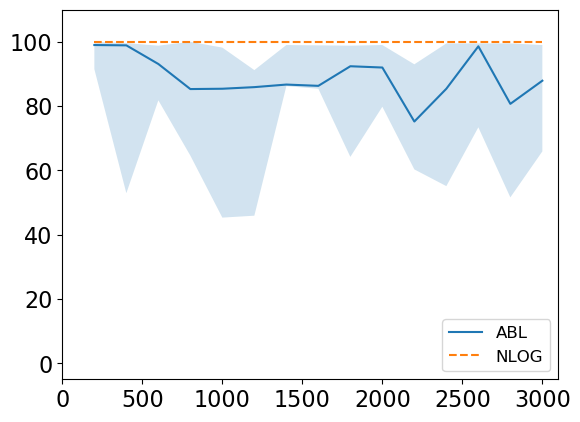} 	& \x\x \includegraphics[width=0.24\textwidth]{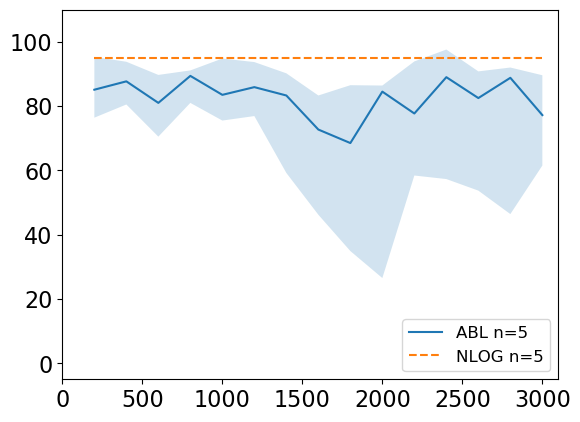}
& \x\x \includegraphics[width=0.24\textwidth]{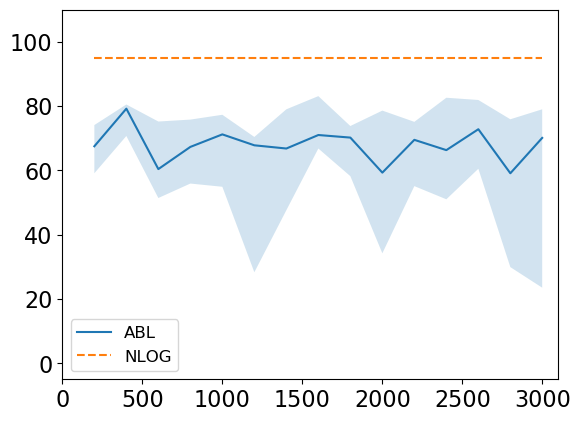} \\
\x
& \x\x $\mat(5)$
& \x\x $\memberof(n)$
& \x\x $\minichessBSV(n)$
& \x\x $\minichessISK(n)$ \\[-0.25ex]
\x \rotatebox{90}{\hspace{0.2cm}{\footnotesize accuracy \% / iter.}} 	 	
& \x\x \includegraphics[width=0.24\textwidth]{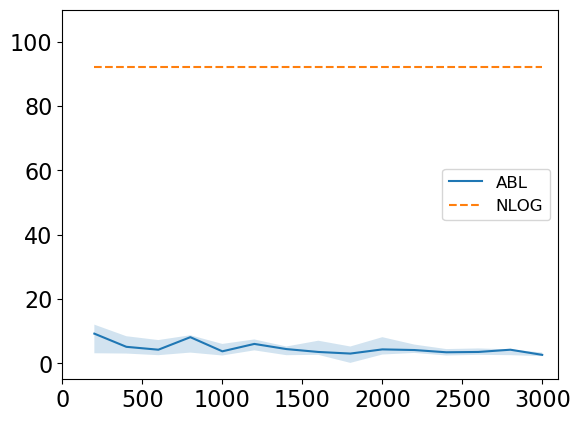} 
& \x\x \includegraphics[width=0.24\textwidth]{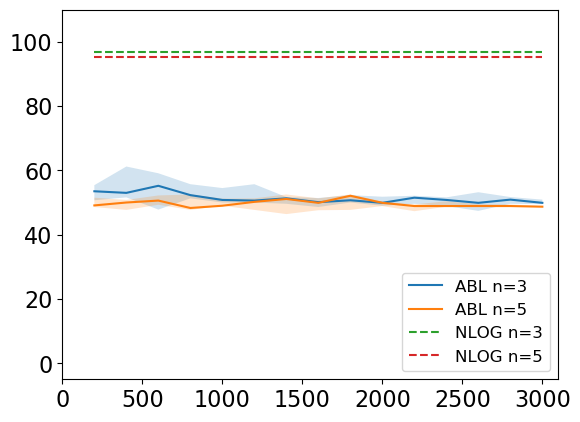} 	
& \x\x \includegraphics[width=0.24\textwidth]{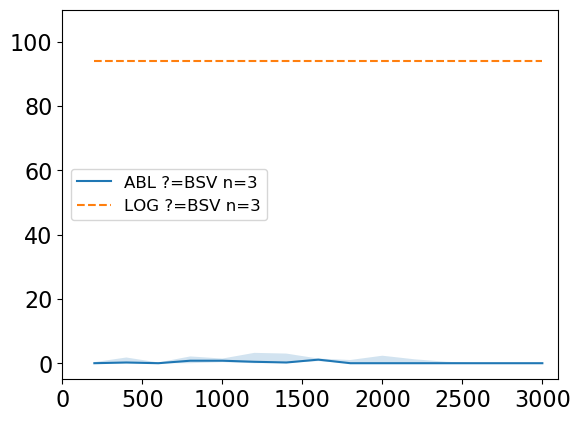} 
& \x\x \includegraphics[width=0.24\textwidth]{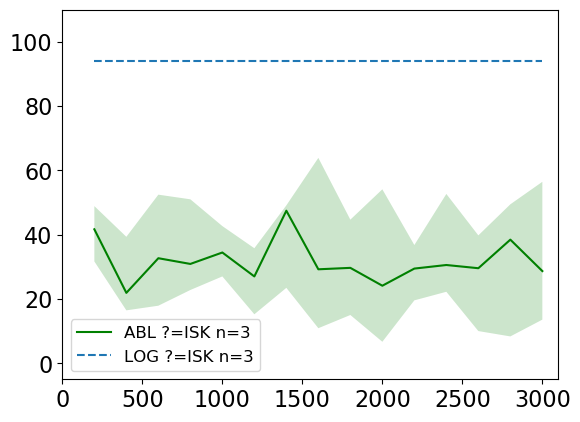}
\end{tabular}\\[5pt]

\caption{Empirical results for $\abl$. Solid lines and lightly-colored areas show, respectively, the average behavior and the variability of the behavior across different repetitions. Dashed lines show the final average accuracy of $\ours$ as reported in Figure~\ref{Figure: Empirical Results 1}. Due to the different training regimes of $\abl$ and $\ours$, an iteration-by-iteration comparison is not meaningful.}
\label{Figure: Empirical Results 2}
\end{figure*}

\begin{figure}
\begin{tabular}[t]{cc}
\x
\x\x \includegraphics[width=0.24\textwidth]{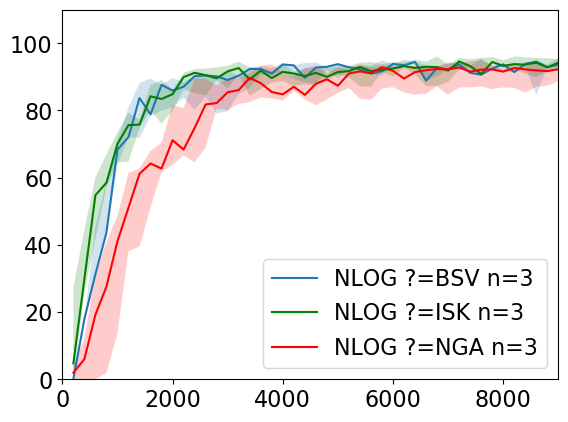} & \x\x \includegraphics[width=0.234\textwidth]{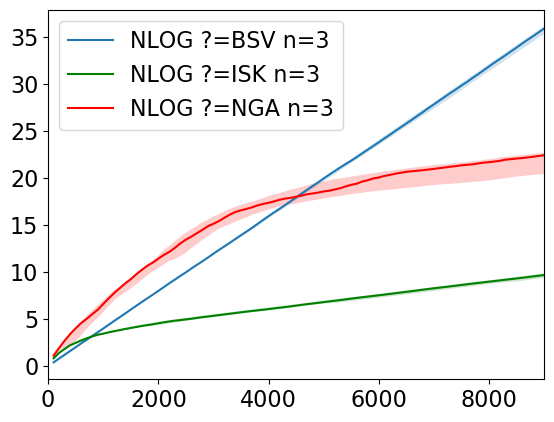} 
\end{tabular} 
\caption{Test accuracy percentage (left) and training time in minutes (right) versus number of iterations for $\minichess(n)$.}\label{figure:minichess}
\end{figure}

We have empirically assessed the training time and test accuracy of our proposed compositional framework, hereafter abbreviated as $\ours$, against three prior approaches that share the same goals with us: $\deepproblog$ \cite{DBLP:journals/corr/abs-1907-08194}, $\neurasp$ \cite{neurasp} and $\abl$ \cite{ABL}. Comparing with other architectures, such as \cite{pmlr-v70-gaunt17a}, which are concerned not only with neural-module learning, but also with symbolic-module learning, is beyond the scope of the current paper. 

\subsection{Implementation} %
\label{Subsection: Implementation}

Abductive feedback in $\ours$ was computed using the A-system \cite{10.1007/3-540-45402-0_29} running over SICStus Prolog 4.5.1. Each abductive feedback $\varphi$ was grounded (and, hence, effectively propositional) by construction, which facilitated the use of semantic loss \cite{pmlr-v80-xu18h} for training the neural module. The semantic loss of $\varphi$ was computed by treating each atom in $\varphi$ as a Boolean variable, weighted by the activation value of the corresponding output neuron of the neural module, and by taking the negative logarithm of its \emph{weighted model count} (WMC)  \cite{CHAVIRA2008772}. For the purposes of computing WMC, $\varphi$ was first compiled into an arithmetic circuit \cite{10.5555/2283516.2283536}. 

In order to avoid recomputing the same models or the same abductive feedbacks during training, we used caching across all the systems that we evaluated. Furthermore, we encoded the theories of the symbolic modules with an eye towards minimizing the time to perform abduction, grounding, or inference. Experiments were ran on an Ubuntu 16.04 Linux PC with Intel i7 64-bit CPU and 94.1 GiB RAM.

\subsection{Data and scenarios} %
\label{Subsection: Data and scenarios}

Benchmark datasets have been used to provide inputs to the neural module as follows: MNIST~\cite{lecun-gradientbased-learning-applied-1998} for images of digits; HASY~\cite{DBLP:journals/corr/Thoma17} for images of math operators; GTSRB~\cite{Stallkamp-IJCNN-2011} for images of road signs.
Below we describe each experimental scenario:

\begin{description}

\item[{$\add$}] \cite{pmlr-v70-gaunt17a}: The input is a ${2 \times 2}$ grid of images of digits. The output is the four sums of the pairs of digits in each row / column. The symbolic module computes the sum of pairs of digits.

\item[{$\operator$}] (new; $\add$ with program induction): The input is a ${2 \times 2}$ grid of images of digits. The output is the four results of applying the math operator $op$ on the pairs of digits in each row / column. The math operator $op$ in $\set{+,-,\times}$ is fixed for each row / column but \emph{unknown}. The symbolic module computes the sum, difference, and product of pairs of digits. The neural module seeks to induce the unknown operator and to recognize the digits.

\item[{$\apply$}] \cite{pmlr-v70-gaunt17a}: The input is three digits $d_1, d_2, d_3$ and a ${2 \times 2}$ grid of images of math operators $op_{i,j}$. The output is the four results of applying the math operators in each row / column on the three digits (e.g., ${d_1~op_{11}~ d_2~ op_{12}~ d_3}$). The symbolic module computes results of applying pairs of math operators on three digits.

\item[{$\dba(n)$}] \cite{ABL}: The input is a mathematical expression comprising $n$ images of \set{0,1} digits and math operators (including the equality operator). The output is a truth value indicating whether the mathematical expression is a valid equation. The symbolic module evaluates the validity of an equation. Our $\dba$ scenario extends that from \cite{ABL} by allowing math operators to appear on both sides of the equality sign.

\item[{$\mat(n)$}] \cite{pmlr-v70-gaunt17a}: The input is a mathematical expression comprising $n$ images of digits and math operators. The output is the result of evaluating the mathematical expression. The symbolic module computes results of math operators on integers.

\item[{$\pat(n)$}] \cite{pmlr-v70-gaunt17a}: The input is an ${n \times n}$ grid of images of road signs and two symbolically-represented grid coordinates. The output is a truth value indicating whether there exists a path from the first to the second coordinate. The symbolic module determines valid paths between coordinates given as facts.

\item[{$\memberof(n)$}] (new): The input is a set of $n$ images of digits and a single symbolically-represented digit. The output is a truth value indicating whether the single digit appears in the set of digits. The symbolic module determines set membership of an element given as a fact.
\end{description}

We have, further, used the chess domain from our running example to highlight certain (new) features of our framework: \emph{a richer class of theories}, \emph{non-declarative theories}, and \emph{neural-guided abduction}. We denote by $\textbf{\minichessBSV}(n)$ and $\textbf{\minichessNGA}(n)$, the scenarios corresponding, respectively, to Example~\ref{example:chess4} and Example~\ref{example:chess6}: in the former scenario, the full abductive feedback is used to train the neural module, and in the latter scenario a \textbf{non-declarative} theory is used to enumerate and evaluate, against the neural predictions, the various abductive proofs to select which parts of the abductive feedback to retain. We also consider a third variant that sits between the former two, called $\textbf{\minichessISK}(n)$, which roughly corresponds to Example~\ref{example:chess5}, but rather than receiving the positions of the three pieces from a confident neural module, it receives them as externally-provided (and noiseless) information. In all scenarios, the chess pieces are represented by images of digits.

\subsection{Results and Analysis} %
\label{Subsection: Results and Analysis}

Results of our empirical evaluation are shown in Table~\ref{Table: Empirical Results}, and in Figures~\ref{Figure: Empirical Results 1}, \ref{Figure: Empirical Results 2}, and \ref{figure:minichess}. Each system was trained on a training set of $3000$ samples, and was ran independently 10 times per scenario to account for the random initialization of the neural module or other system stochasticity. Training was performed over $3$ epochs for $\ours$, $\deepproblog$ and $\neurasp$, while the training loop of $\abl$ was invoked $3000$ times. Note that there is no one-to-one correspondence between the training loop of $\abl$ and that of the other three systems: in each iteration, $\abl$ considers multiple training samples and based on them it trains the neural component multiple times. In all systems, the neural module was trained using the Adam algorithm with a learning rate of $0.001$. 

Results on running $\deepproblog$ on the $\minichess(n)$ suite of scenarios are not available, since $\deepproblog$'s syntax does not readily support the integrity constraints (nor the procedural constructs for the $\minichessNGA(n)$ scenario) in the symbolic module. Since neural-guided abduction is not supported by any of the other three systems, we report results on $\minichessNGA$ only for $\ours$. 

The results offer support for the following conclusions:

\cond{C1} The average accuracy of $\ours$ is comparable to, or better than, that of the other systems. $\ours$ performs similarly to $\deepproblog$ on those scenarios that are supported by the latter and in which $\deepproblog$ does not time out, while it may perform considerably better than $\neurasp$ and $\abl$. For example, the average accuracy of $\ours$ is up to $70\%$ higher than that of $\neurasp$ in the $\mat$ scenarios, and up to $40\%$ higher than that of $\neurasp$ in the $\memberof$ scenarios.  

$\neurasp$ and $\abl$ are vulnerable to weak supervision signals, as their performance decreases when the number of abductive proofs per training sample increases. For example, the average accuracy of $\abl$ drops from $69.7\%$ in ${\mat(3)}$ to $6.1\%$ in ${\mat(5)}$, while it drops from $44\%$ in $\minichessISK(n)$, where each training sample is provided with the coordinates of the non-empty cells, to less than $1\%$ in $\minichessBSV(n)$ where no such information is provided.

With regards to $\abl$, this phenomenon may be attributed to the consideration of a \emph{single} abductive proof per training sample instead of considering all the relevant abductive proofs as $\ours$ does. Considering a single abductive proof may result in excluding the correct one; i.e., the one corresponding to the true state of the sample input. Notice that when the number of abductive proofs per training sample increases, the probability of excluding the right abductive proof from consideration increases as well, resulting in very weak supervision signals, as seen in $\minichessBSV$.

\cond{C2} Compared to $\neurasp$ and $\abl$, $\ours$ is less sensitive to the initialization of the neural module. For example, the accuracy of $\neurasp$ spans $15\%$--$94\%$ in $\mat(3)$, and that of $\abl$ spans $57\%$--$94\%$ in $\apply$.

With regards to $\abl$, this sensitivity may be, again, attributed to the consideration of a \emph{single} abductive proof per training sample. The learning process of $\abl$ obscures and abduces part of the neural predictions, so that the modified predictions are consistent with the theory and also lead to the entailment of the sample label (see the ``Related Work” section). Considering a single abductive proof has high chances of missing the right one, and hence the training process ends up being biased on the obscuring process, which, in turn, depends upon the initial weights of the neural module. 

\cond{C3} The average training time of $\ours$ may be significantly less than that of the other systems. For example, the average total training time is: 16m47s for $\ours$ in $\mat(5)$ versus 22m48s for $\deepproblog$ in the simpler $\mat(3)$ scenario; 42m93s for $\ours$ in $\pat(6)$ versus $\deepproblog$ and $\neurasp$ timing out in the simpler $\pat(4)$ scenario; 16m for $\ours$ in $\minichessISK(3)$ versus 125m for $\abl$ in the same scenario.

With regards to $\abl$, its high training time may be attributed to its trial-and-error use of abduction. At each training iteration, an optimization process obscures and performs abduction \emph{multiple times} over different subsets of the training samples. It holds, in particular, that although $\abl$ computes a single abductive proof per training sample, it may perform abduction multiple times for the same sample. 

With regards to $\neurasp$, its high training time may be attributed to the grounding that $\neurasp$ applies on the theory; i.e., computing all the consequences that are semantically entailed. Instead of computing all such forward-reasoning consequences, abduction is driven by the sample label, and evaluates (and grounds) only the relevant part of the theory. It is worth noting, however, that $\neurasp$ achieves comparable accuracy to $\ours$ in less training time in the $\add$ and $\dba$ scenarios. Its training time is also lower than that of $\ours$ in the two $\mat(n)$ scenarios, however, for these cases its accuracy is very poor. 

\cond{C4} When compared to $\minichessBSV(3)$, the use of side-information in $\minichessISK(3)$ and $\minichessNGA(3)$ leads to asymptotically faster training. The higher training time during the earlier iterations, which is particularly pronounced in the $\minichessNGA(3)$ scenario (see Figure~\ref{figure:minichess}), corresponds to the phase where new abductive proofs are still being computed. Recall that in $\minichessBSV(3)$ an abductive proof is distinct for each label (i.e., mate, draw, safe), whereas in $\minichessISK(3)$ and  $\minichessNGA(3)$ an abductive proof is distinct for each combination of label and side-information. Once the bulk of the distinct abductive proofs is computed and cached, the training time per iteration drops. Unsurprisingly, this initial phase is longer for the $\minichessNGA(3)$ scenario, where the distinct abductive proofs are more, as they depend on a more variable space of side-information. 

The average end accuracy for the $\minichess(3)$ scenarios is comparable; see Table~\ref{Table: Empirical Results}. The average \emph{interim} accuracy of $\minichessNGA(3)$ is, however, relatively lower during early training, where the neural module predictions are still highly noisy / random. Specifically, the average accuracy at $1000$ iterations is: ${73.9 \pm 1.5}$ for $\minichessBSV(3)$, ${73.4 \pm 5.2}$ for $\minichessISK(3)$, ${51.1 \pm 7.9}$ for $\minichessNGA(3)$. 

\textbf{Scalability:} Computing abductive proofs is intractable (NP-hard to decide their existence; \#P-hard to enumerate/count them). Neural-guided abduction reduces this cost in practice by excluding irrelevant proofs, but the problem remains worst-case hard. However, since abductive proofs are a function of only the sample label and side-information, $\ours$ can cache and reuse them across different training samples, showing that in 
practice our approach can be more computationally efficient than prior art, e.g., $\abl$.

\section{Related Work} %
\label{Section: Related Work}

Although $\abl$ shares with $\ours$ the high-level idea of using abduction, it does so by employing an ad hoc optimization procedure. In each training iteration over a given set ${\set{\tuple{x^j,f(x^j)}}_j}$ of training samples, $\abl$ first considers different subsets $S^t$ of the training set and performs the following steps: 
\cond{i} it gets the neural predictions for each element in $S^t$, \cond{ii} it obscures a subset of the neural predictions (both within the same and across different training samples), and \cond{iii} it abduces the obscured predictions so that the resulting predictions are consistent with the background knowledge. Let $S^{*}$ be the largest $S^t$ satisfying the theory after obscuring and abducing. For each ${s_i \in S^{*}}$, $\abl$ trains multiple times the neural component using obscured and abduced neural predictions. As our empirical results show, the optimization procedure applied to obscure and abduce the neural predictions may be time-consuming and ineffective, even though a single abductive proof is computed each time.

The second system which shares $\ours$'s objectives is $\deepproblog$ \cite{DBLP:journals/corr/abs-1907-08194}, which works by reducing the problem of learning and inference of a neural-symbolic system to the problem of learning and inference over a probabilistic logic program. The reduction works by treating the outputs of the neural module as probabilistic facts. The accuracy of $\deepproblog$ is comparable to that of $\ours$, but it supports fewer semantic constructs (e.g., integrity constraints) and it requires significantly more training time. $\neurasp$ shares the same high-level approach with $\deepproblog$, but reduces, instead, to the more expressive probabilistic answer set programs (ASP) \cite{neurasp}. As supported by our empirical evidence, its performance may be lower than that of $\ours$. Furthermore, using an ASP solver may be computationally expensive, since it involves computing all the consequences that are semantically entailed by the theory. Instead, our training approach is goal-driven, since the computation of the abductive feedback involves evaluating only the part of the theory that relates to the training label.

Our work can be seen as an extension to \cite{pmlr-v80-xu18h}, where a given \emph{fixed} propositional formula is used to define a loss function based on weighted model counting, and this loss function is used, in turn, to regulate the training of the neural module. In contrast to \cite{pmlr-v80-xu18h}, our work computes a sample-specific formula to regulate the training of the neural component based on the label of each sample.

\textbf{Broader area of neural-symbolic integration:} The work in~\cite{DBLP:conf/iclr/ParisottoMS0ZK17,kalyan2018neuralguided,DBLP:conf/iclr/BalogGBNT17} uses ML to help perform faster and more data efficient program induction, while saving from designing heuristics. To tackle rule induction in the presence of noisy examples, the work in~\cite{DBLP:journals/jair/EvansG18} reduces inductive logic programming to a problem of minimizing a differentiable loss. Other frameworks that deal with rule induction under noisy data are Neural Logic Programming \cite{DBLP:conf/nips/YangYC17} and DRUM \cite{DBLP:conf/nips/SadeghianADW19}. Neural Theorem Prover \cite{rocktaschel2017ntp} is an alternative to Prolog’s QA engine to support noisy theories. It proceeds by  embedding predicates and constants into a vector space and uses vector distance measures to compare them. Neural Logic Machines \cite{DBLP:conf/iclr/DongMLWLZ19} implements rules inside a tensor network providing thus the ability to reason uniformly over neural modules and logical theories. However, its semantics is not connected to any logic semantics (e.g., Tarski, Sato, or fuzzy) and no soft or hard constraints are imposed at inference time. 

\section{Conclusion} %
\label{Section: Conclusion}

This work has introduced a compositional framework for neural-symbolic integration that utilizes abduction to support the uniform treatment of symbolic modules with theories beyond any given logic or a declarative representation altogether. Our empirical results have  demonstrated not only the practical feasibility of this perspective, but also that it can outperform various state-of-the-art approaches in terms of cross-domain applicability, testing accuracy, and training speed. Two are the key directions for future work: \cond{i} further consideration of non-logic or non-declarative theories for the symbolic module; \cond{ii} explicit treatment of symbolic-module learning, which, unlike program induction, will not delegate the burden of learning to the neural module.

\section*{Acknowledgements}

This work was supported by funding from the EU's Horizon 2020 Research and Innovation Programme under grant agreements no.\ 739578 and no.\ 823783, and from the Government of the Republic of Cyprus through the Directorate General for European Programmes, Coordination, and Development. The authors would like to thank Timothy Hospedales for early discussions in the general area of neural-symbolic integration, and Antonis Kakas for help with the abduction system that was used in this work.

\bibliography{references}

\end{document}